\documentclass{article}

\usepackage{microtype}
\usepackage{graphicx}
\usepackage{subcaption}
\usepackage{booktabs}
\usepackage{hyperref}
\usepackage{dblfloatfix}
\usepackage{float}
\usepackage{placeins}

\usepackage{algorithm}
\usepackage{algorithmic}
\usepackage{makecell} 

\usepackage[preprint]{icml2026}

\definecolor{icmlciteblue}{RGB}{0,102,204}
\hypersetup{colorlinks=true,linkcolor=icmlciteblue,citecolor=icmlciteblue,urlcolor=icmlciteblue}

\renewenvironment{abstract}
{%
  \centerline{\large\bf Abstract}
  \vspace{-0.12in}\begin{list}{}{
    \setlength{\leftmargin}{0.6em}
    \setlength{\rightmargin}{0.6em}
  }\item\relax\itshape
}{\end{list}\vskip 0.12in}

\usepackage{amsmath}
\usepackage{amssymb}
\usepackage{mathtools}
\usepackage{amsthm}
\usepackage[capitalize,noabbrev]{cleveref}
\usepackage{bm}

\theoremstyle{plain}

\theoremstyle{definition}

\theoremstyle{remark}

\icmltitlerunning{Single-Layer Transformer One-Step Generation}

\begin{document}

\twocolumn[
  \icmltitle{A one-step generation model with a Single-Layer Transformer: Layer number re-distillation of FreeFlow}

  \icmlsetsymbol{equal}{*}

  \begin{icmlauthorlist}
    \icmlauthor{Haonan Wei}{ieu}
    \icmlauthor{Linyuan Wang}{ieu}
    \icmlauthor{Nuolin Sun}{ieu}
    \icmlauthor{Zhizhong Zheng}{ieu}
    \icmlauthor{Lei Li}{ieu}
    \icmlauthor{Bin Yan}{ieu}
  \end{icmlauthorlist}

  \icmlaffiliation{ieu}{Information Engineering University}

  \icmlcorrespondingauthor{Lei Li}{leehotline@163.com}

  \icmlkeywords{keyword1, keyword2}

  \vskip 0.3in

  \begin{center}
  \begingroup
  \resizebox{\textwidth}{!}{\includegraphics{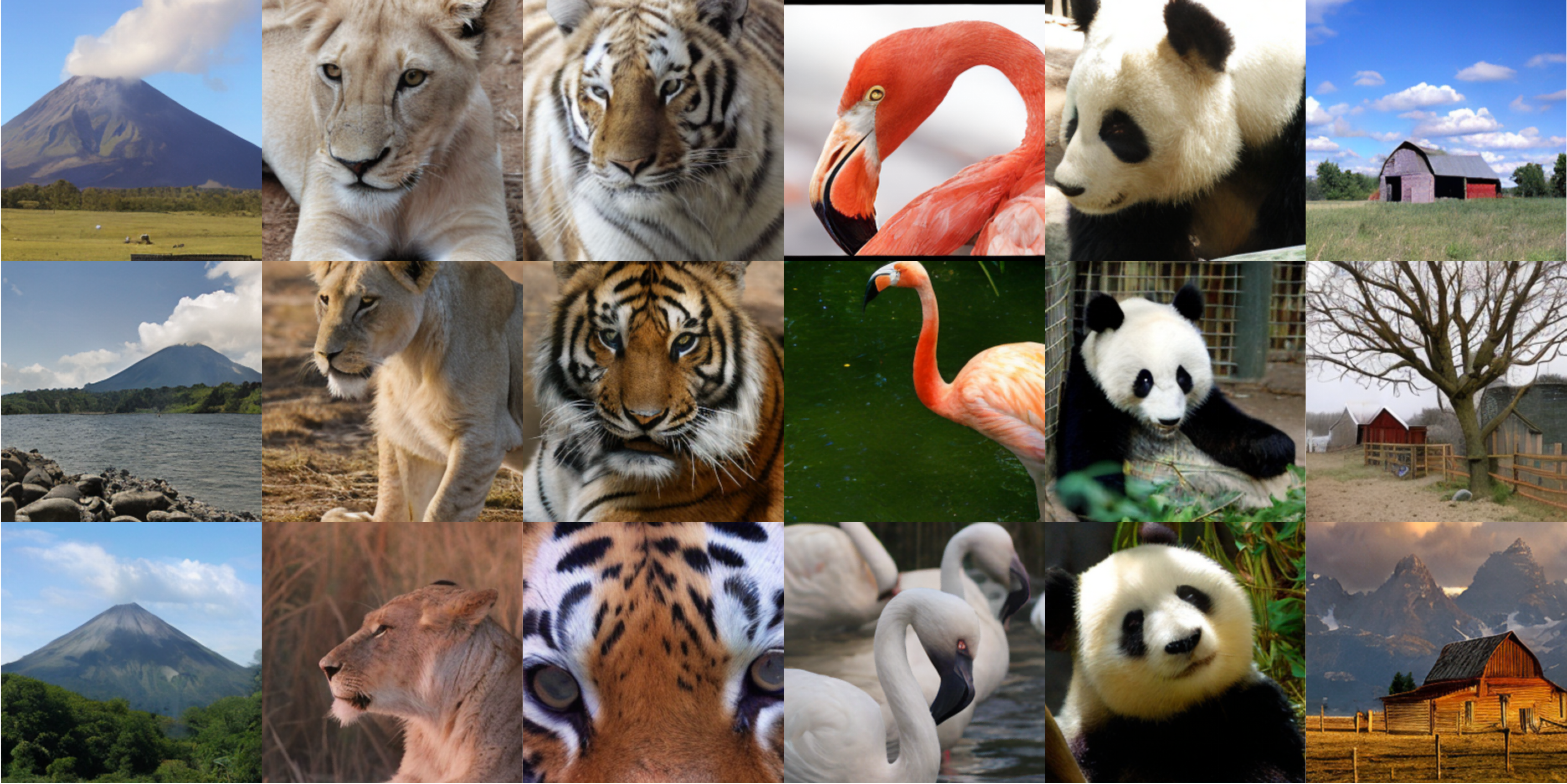}}
  \captionof{figure}{Samples generated by our Scout-and-Refine strategy on ImageNet 256$\times$256. Each image is produced by first using SLT to evaluate 100 candidate noises, selecting the best one, and then generating the final image with FreeFlow. The results demonstrate that our method can produce high-quality, diverse samples across various categories, including animals, objects, and scenes.}
  \label{fig:samples}
  \endgroup
  \end{center}
]

\printAffiliationsAndNotice{}

\begin{abstract}
Currently, Flow matching methods aim to compress the iterative generation process of diffusion models into few or even a single step, with MeanFlow and FreeFlow being representative achievements of one-step generation based on Ordinary Differential Equations (ODEs). We observe that the 28-layer Transformer architecture of FreeFlow can be characterized as an Euler discretization scheme for an ODE along the depth axis, where the layer index serves as the discrete time step. Therefore, we distill the number of layers of the FreeFlow model, following the same derivation logic as FreeFlow, and propose SLT (Single-Layer Transformer), which uses a single shared DiT block to approximate the depth-wise feature evolution of the 28-layer teacher.  During training, it matches the teacher’s intermediate features at several depth patches, fuses those patch-level representations, and simultaneously aligns the teacher’s final velocity prediction. Through distillation training, we compress the 28 independent Transformer Blocks of the teacher model DiT-XL/2 into a SLT with only one shared Transformer Block, reducing the parameter count from 675M to 4.3M. Furthermore, leveraging its minimal parameters and rapid sampling speed, SLT can screen more candidate points in the noise space within the same timeframe, thereby selecting higher‑quality initial points for the teacher model FreeFlow and ultimately enhancing the quality of generated images. Experiments results demonstrate that within a time budget comparable to two random samplings of the teacher model, our method performs over 100 noise screenings and produces a high-quality sample through the teacher model using the selected points. Quality fluctuations caused by low-quality initial noise under a limited number of FreeFlow sampling calls are effectively avoided, substantially improving the stability and average generation quality of one-step generation.
\end{abstract}

\section{Introduction}
Diffusion models have become the mainstream paradigm in image generation, replacing GANs~\citep{goodfellow2014gan} due to their excellent generation quality and diversity. However, high-quality generation comes at the cost of high computational complexity. Traditional diffusion model sampling requires numerically solving Ordinary Differential Equations(ODE), a process that typically involves dozens or even hundreds of iterative solver steps, with each step requiring a complete neural network forward pass. Many widely used samplers can be described in a unified SDE/ODE formulation, which clarifies that sampling essentially reduces to repeated model evaluations and thus remains expensive~\citep{song2021sde}. For example, standard sampling in Denoising Diffusion Probabilistic Models(DDPM)~\citep{ho2020ddpm} requires 1000 steps. Even with acceleration methods like DDIM~\citep{song2020ddim}, 20--50 steps are still needed to achieve satisfactory generation quality, and modern samplers and training formulations further improve efficiency and quality~\citep{karras2022edm,dhariwal2021beatgans,nichol2021improvedddpm,lu2022dpmsolver}. In addition to advances in numerical solvers, distillation-based methods and one-step (or few-step) generation paradigms have emerged as prominent approaches for reducing sampling costs~\citep{salimans2022progressive,song2023consistency,liu2022rectifiedflow}.

To bypass the bottleneck of iterative solving, Flow Matching \citep{lipman2022flowmatching} aims to directly learn the integral of the underlying probability flow ODEs, enabling one-step generation. Within this framework, Flow Map methods \citep{geng2025meanflow, tong2025freeflow} propose to parameterize the solution directly. Specifically, MeanFlow \citep{geng2025meanflow} introduces a key parameterization where a network $\text{F}_{\theta}$ represents the average velocity along the trajectory: $\text{f}_{\theta}(\text{x}_{\text{t}},t,s)=\text{x}_{\text{t}}+(t-s)\cdot\text{F}_{\theta}(\text{x}_{\text{t}},t,s)$. Here, $\text{f}_{\theta}$ directly approximates the flow map from time $t$ to $s$, allowing the model to compress integration steps by learning this averaged field. Following this formulation, FreeFlow \citep{tong2025freeflow} also achieves one-step sampling. Furthermore, it derives a prior-anchored training objective that eliminates reliance on a specific data noising process and mitigates the teacher-student gap.

Drawing on FreeFlow's distillation method for integration steps, we revisit the Transformer structure. The residual stacking in Transformers can be interpreted through an ODE viewpoint, each residual block corresponds to one Euler discretization step. After the same derivation as FreeFlow, the feature transformation of each layer in Transformer can be expressed as an instantaneous velocity field, and thus expressed as average velocity field. Through targeted distillation, we compress the 28-layer Transformer of the FreeFlow teacher model into a Single-Layer Transformer (SLT) that shares one block across depth, achieving a drastic reduction in parameters. Moreover, leveraging SLT's extremely low inference cost, we design a Scout-and-Refine strategy that uses SLT to rapidly screen high-quality initial points in the noise space, which are then passed to FreeFlow for final generation. The main contributions are summarized as follows:
\begin{enumerate}
\item We interpret the 28-layer Transformer in FreeFlow as 28 Euler discretization steps of an ODE solver. Following FreeFlow's flow-map distillation formulation, we distill these 28 depth steps into a single shared-weight DiT block, reducing parameters from 675M to 4.3M.
\item We introduce an efficient noise screening strategy that leverages the minimal inference cost of SLT to perform Best-of-$N$ selection in the noise space. Experiments show that within the time required for 2 samplings of FreeFlow, over 100 noise screenings can be evaluated and one high-quality image can be generated by the teacher model FreeFlow. This enables high-quality generation with only a limited number of teacher sampling calls, outperforming random sampling under the same budget.

\end{enumerate}

\section{Method}
Building upon FreeFlow's one-step distillation framework in Section~\ref{sec:freeflow-time}, we first reformulate the depth of a Transformer as a discretized ODE process. This reformulation allows us to derive a depth-wise distillation objective, culminating in the proposal of our Single-Layer Transformer (SLT) in Section~\ref{sec:depth-distill}. Capitalizing on SLT's efficiency, we then present the Scout-and-Refine sampling strategy in Section~\ref{sec:scout-refine}, which completes our method.

\subsection{FreeFlow: Flow Map Distillation in the Temporal Dimension}
\label{sec:freeflow-time}
FreeFlow models the generation process of diffusion models as a continuous flow from the noise distribution to the data distribution. Define the time variable $t\in[0,1]$, and the state trajectory $\bm{x}_t$ satisfies the ODE:
\begin{equation}
\frac{d\bm{x}(t)}{dt} = -\bm{u}(\bm{x}(t), t)
\end{equation}
where $\bm{u}(\bm{x},t)$ is the instantaneous velocity field. Generation requires reverse integration along this ODE, i.e., calculating the integral $\phi_u(\bm{x}_t,t,s)=\bm{x}_t+\int_{s}^{t}\bm{u}(\bm{x}(\tau),\tau)d\tau$. Due to the complex curvature of the trajectory, numerical integration often requires a large number of NFEs (Number of Function Evaluations).

To accelerate sampling, the core idea of FreeFlow is to learn a mean velocity field $F_\theta$, such that $f_\theta$ approximates $\phi_u$. Specifically, the integral of the instantaneous velocity over the interval $[1-\delta,1]$ is averaged to obtain the mean velocity required for one-step generation:
\begin{equation}
\label{eq:2}
F_\theta(\bm{z},\delta)\approx\frac{1}{\delta}\int_{1}^{1-\delta}-\bm{u}(\bm{x}(\tau),\tau)d\tau
\end{equation}
The MeanFlow approach often obtains consistency constraints by perturbing the starting time $t$. However, in the data-free setting, the starting point is fixed at $t=1$ with $x_t=z$, making perturbing the starting point meaningless. FreeFlow thus adopts a symmetric strategy, perturbing the end point (equivalent to differentiating with respect to $\delta$). Under this reparameterization, the key identity is derived:
\begin{equation}
F_{\theta^*}(\bm{z},\delta)+\delta\frac{\partial}{\partial\delta}F_{\theta^*}(\bm{z},\delta)=\bm{u}(f_{\theta^*}(\bm{z},\delta),1-\delta)
\end{equation}
where $f_\theta(\bm{z},\delta)=\bm{z}-\delta F_\theta(\bm{z},\delta)$ is the state predicted by the model. Based on this, a training objective that only requires sampling from the prior is constructed:
\begin{equation}
L_{FreeFlow}=E_{\bm{z},\delta}\left\| F_\theta(\bm{z},\delta)-sg(\bm{u}_{target}) \right\|^2
\end{equation}
where $\bm{u}_{target}=\bm{u}(f_\theta(\bm{z},\delta),1-\delta)-\delta\frac{\partial}{\partial\delta}F_\theta(\bm{z},\delta)$, and $sg(\cdot)$ denotes stop gradient.

\subsection{Flow Map Distillation in the Depth Dimension}
\label{sec:depth-distill}
\subsubsection{Depth ODE Formulation}
Inspired by FreeFlow's formulation of the generation process as a residual structure in the temporal dimension, we extend its residual-based modeling concept from the temporal dimension to the depth dimension of the network. We observe the 28-layer Transformer structure of FreeFlow, where each layer can be expressed in a residual form as follows:
\begin{equation}
\mathbf{h}^{(l)} = \mathbf{h}^{(l - 1)} + F(\mathbf{h}^{(l - 1)},l),\quad l = 1,\ldots,L
\end{equation}
where $F$ denotes the transformation function at layer $l$, which consists of layer normalization (LN), self-attention (SelfAttn), and feed-forward network (FFN). The complete form is given by:
\begin{equation}
\begin{aligned}
F(\mathbf{h}^{(l-1)},l)=FFN\Big( LN\big( \mathbf{h}^{(l-1)} \\
\qquad + SelfAttn( LN( \mathbf{h}^{(l-1)} ),l ) \big),l \Big)
\end{aligned}
\end{equation}

The additive form $\mathbf{h}^{(l)}=\mathbf{h}^{(l-1)}+F(\mathbf{h}^{(l-1)},l)$ explicitly corresponds to the residual connection. Specifically, the identity branch carries the previous state $\mathbf{h}^{(l-1)}$, while $F$ provides the residual increment that updates the features. The above process can be rewritten as the discrete form of a continuous depth ODE:
\begin{equation}
\frac{d\mathbf{x}(t)}{dt}=-\mathbf{u}(\mathbf{x}(t),t)
\end{equation}
Through the same derivation process as FreeFlow, we obtain:

\begin{equation}
\label{eq:8}
\mathbf{G}_\phi\big(\mathbf{h}, \delta\big) \approx \frac{1}{\delta} \int_{1}^{1-\delta} -\mathbf{u}(\mathbf{h}(\tau), \tau) \, d\tau
\end{equation}

Combining the above, the output can be directly computed from the input via:
\begin{equation}
\mathbf{h}(1-\delta) \approx \mathbf{h}(1) - \delta \, \mathbf{G}_\phi\big(\mathbf{h}(0), \delta\big)
\end{equation}

\textbf{Depth-dimension flow map.} Analogous to FreeFlow's time-dimension flow map, we define a depth-dimension counterpart that compresses the entire backbone $(\tau: 0 \to 1)$ into a single transformation. With diffusion time fixed at $t=1$ and $\mathbf{z} \sim \mathcal{N}(0, \mathbf{I})$, the learned transformation field satisfies:
\begin{equation}
\mathbf{G}_\phi\big(\mathbf{h}(0))\;\approx\;\int_{0}^{1}\mathbf{u}\big(\mathbf{h}(\tau),\tau)\,d\tau
\end{equation}

In particular, for the full depth interval $(\delta = 1)$, we can obtain the average velocity field over the interval $[0,1]$:
\begin{equation}
\mathbf{h}(1)\;\approx\;\mathbf{h}(0)+\mathbf{G}_\phi\big(\mathbf{h}(0))
\end{equation}

\subsubsection{Distillation Process}
A comparison between Eq.~\eqref{eq:2} and Eq~\eqref{eq:8} reveals that the $L$-layer forward propagation of Transformer can be interpreted as a discretization of an ODE along the depth dimension, where the number of layers $L$ corresponds to the number of discrete time steps. This equivalence allows us to directly adopt FreeFlow's flow-map distillation to achieve network depth compression.

If we introduce the normalized depth coordinate $\tau=l/L\in[0,1]$ and regard the inter-layer transformation asan Euler discretization step with step size $\Delta\tau=1/L$,then the 28-layer Transformer structure of FreeFlow can be expressed as $\Delta\tau=1/28$. Based on this formulation, we propose the Single-Layer Transformer (SLT). Its core idea is that since the continuous evolution along the depth dimension can be fully described by a depth-averaged transformation field $\mathbf{G}_\phi$, this field can be parameterized using a single, shared Transformer block, thereby compressing the multi-layer architecture into one layer. SLT contains only one DiT Block with shared weights, which is conceptually related to parameter sharing across layers in Transformers~\citep{dehghani2019universal,lan2019albert,bai2019deq}. In our knowledge distillation framework, the SLT model acts as the student, trained to replicate the functionality of the teacher—the original multi-layer Transformer.

To simulate the $L$-layer propagation of the teacher network with this single layer, we unroll it along the depth dimension into $K$ discrete time steps. The depth scalar corresponding to the $k$-th step is:
\begin{equation}
\tau_k=\frac{k}{K-1},\quad k=0,1,\ldots,K-1
\end{equation}
Denote the feature of the student at the $k$-th step as $\hat{\mathbf{h}}_k$, its forward iteration is:
\begin{equation}
\hat{\mathbf{h}}_{k+1}=\hat{\mathbf{h}}_k+G_\phi(\hat{\mathbf{h}}_k,\tau_k),\quad k=0,\ldots,K-2
\end{equation}

To supervise the representational evolution of the student along the depth trajectory, we extract intermediate features from the teacher network as patches supervision, denoted as $h_T$. Concretely, our teacher is a 28-layer Transformer, while the student performs a $K$-step discrete rollout; we therefore align each student step with a set of evenly spaced teacher layers, and use the teacher features at those layers as the supervision targets for the corresponding student steps.

The total loss function consists of two parts:

\textbf{Output Alignment.} We align the final velocity field predictions of the student and teacher:
{\small
\begin{equation}
L_{output} = \mathbb{E}_{\mathbf{z},y,t,w}\left[ \left\| \mathbf{v}_\phi(\mathbf{z},t,y,w) - \mathbf{v}_T(\mathbf{z},t,y,w) \right\|_2^2 \right]
\end{equation}
}
where $\mathbf{v}_\phi$ and $\mathbf{v}_T$ denote the velocity predictions of the student and teacher models, respectively, conditioned on noise $\mathbf{z}$, class label $y$, diffusion time $t$, and CFG weight $w$.

\textbf{Intermediate Patches Alignment.} We supervise the student's depth trajectory by matching intermediate features:
\begin{equation}
L_{p} = \frac{1}{K}\sum_{i=0}^{K-1}\left\| P_i(\mathbf{h}_i^{student}) - \mathbf{h}_{m(i)}^{teacher} \right\|^2
\end{equation}
where $K$ is the number of depth steps, $\mathbf{h}_i^{student}$ is the student's feature at the $i$-th step, $\mathbf{h}_{m(i)}^{teacher}$ is the teacher's feature at the corresponding layer $m(i)$, and $P_i$ is a linear projection that aligns the hidden dimensions when they differ.

The overall training objective combines both terms:
\begin{equation}
L_{total} = L_{output} + \lambda L_{p}
\end{equation}
where $\lambda$ controls the weight of intermediate alignment loss.

\subsection{Best-of-N Selection with Scout Model}
\label{sec:scout-refine}
The output quality of one-step generation models is highly sensitive to the initial noise $z$, such that small differences in random seeds can lead to significant variations in generation results.

We observe that a notable advantage of SLT lies in its extremely low inference cost, which enables efficient exploration of the noise space. Based on this, we position SLT as a lightweight noise screening module, referred to as Scout, to quickly identify high-quality initial points in the noise domain. Furthermore, since SLT contains only about 0.6\% of the teacher model's parameters, a simple "best-of-N" selection strategy can be adopted: generating N candidate noise vectors, scoring them rapidly with SLT, and forwarding only the top-scoring noise to the teacher model for final generation. This screening process effectively transforms inherent noise sensitivity into a controllable selection step, without significantly increasing overall computational cost, just requiring only a small number of teacher model evaluations.

Specifically, we first use SLT to cheaply preview and score many candidate noises, and then only spend one teacher evaluation on refining the single selected noise. This method significantly reduces the randomness of single-step sampling, while keeping the total inference cost comparable to two teacher samplings. The entire procedure is outlined in Algorithm~\ref{alg:scout-refine}.

\begin{algorithm}[H]
\caption{Scout-and-Refine}
\label{alg:scout-refine}
\begin{algorithmic}[1]
\STATE \textbf{Input:} Scout model SLT; teacher model FreeFlow; scoring function $S(\cdot)$; budget $N$; class label $y$.
\STATE Sample noises $\{z_i\}_{i=1}^N \sim \mathcal{N}(0, I)$.
\FOR{$i=1$ \textbf{to} $N$}
\STATE $\mathbf{x}_i \leftarrow f_\theta(\mathbf{z}_i, 1, y)$ \hfill \textit{}
\STATE $s_i \leftarrow S(\hat{\mathbf{x}}_i)$ \hfill \textit{}
\ENDFOR
\STATE $i^* \leftarrow \arg\max_i s_i$.
\STATE $\mathbf{x}_{\text{final}} \leftarrow \mathbf{v}_T(\mathbf{z}_{i^*}, 1, y)$ \hfill \textit{}
\STATE \textbf{Output:} $x_{\text{final}}$.
\end{algorithmic}
\end{algorithm}

\section{Experiments}
This section first introduces the experimental setup, then demonstrates the distillation effect of SLT , and finally focuses on analyzing the role of the SLT-based Scout-and-Refine sampling strategy in improving generation quality and stability.

\subsection{Experimental Setup}
We adopt a Data-Free distillation strategy. During training, noise $z\sim N(0,I)$, class labels $y\sim U(0,999)$, and CFG weights $w\sim U(1.0,2.0)$ are all randomly sampled online. We use the AdamW optimizer with a base learning rate of $3\times 10^{-5}$, combined with a cosine annealing scheduler with linear warmup (warmup steps = 6,000) for 60,000 iterations of updates. The batch size is set to 32. All experiments are conducted on 9 NVIDIA RTX 3090 GPUs.

FreeFlow serves as both the supervisory source during training and as the final generator in the refinement stage at inference time. The SLT student is trained with two complementary targets, which encourages SLT to approximate the evolution of the teacher's depth-wise features and the prediction of the results. Inference, we adopt a Scout-and-Refine procedure: we draw $N$ candidate noises, use SLT to cheaply preview and score them, and then run the teacher once on the single selected noise to produce the final image. Unless otherwise stated, we set $N=100$, which keeps the overall compute within one teacher evaluation. For fair comparison, we match the wall-clock budget to a baseline of two random FreeFlow samplings.

\subsection{Effectiveness of SLT Distillation}
We systematically evaluate the student model SLT, obtained through our knowledge distillation framework that compresses a 28-layer FreeFlow (DiT-XL) Transformer into a single-layer network designed to emulate its behavior. In this process, we introduce a discrete-step rollout mechanism that aligns each forward step of the student with intermediate layers at different depths in the teacher network, enabling it to simulate the feature evolution from shallow to deep layers of the teacher model.

The SLT student model achieves significant compression over the teacher, reducing the hidden size to 384, the number of attention heads to 6, and the total parameters to only 4.34M. This represents an approximately 155× reduction compared to the teacher model, with detailed configurations provided in Table~\ref{tab:model-config}.

\begin{table}[H]
\caption{Model configurations.}
\label{tab:model-config}
\centering
\small
\setlength{\tabcolsep}{4pt}
\renewcommand{\arraystretch}{1.2} 
\begin{tabular}{cccc}
\toprule
Model & \makecell[c]{Hidden\\Size} & \makecell[c]{Heads} & \makecell[c]{Params\\(M)} \\
\midrule
Teacher (DiT-XL) & 1152 & 16 & 675.00 \\
SLT & 384 & 6 & 4.34 \\
\bottomrule
\end{tabular}
\end{table}

\subsection{Scout-and-Refine Analysis}
\begin{figure*}[t]
\vspace*{-0.8cm}
\centering
\includegraphics[width=\textwidth,height=0.41\textheight,keepaspectratio]{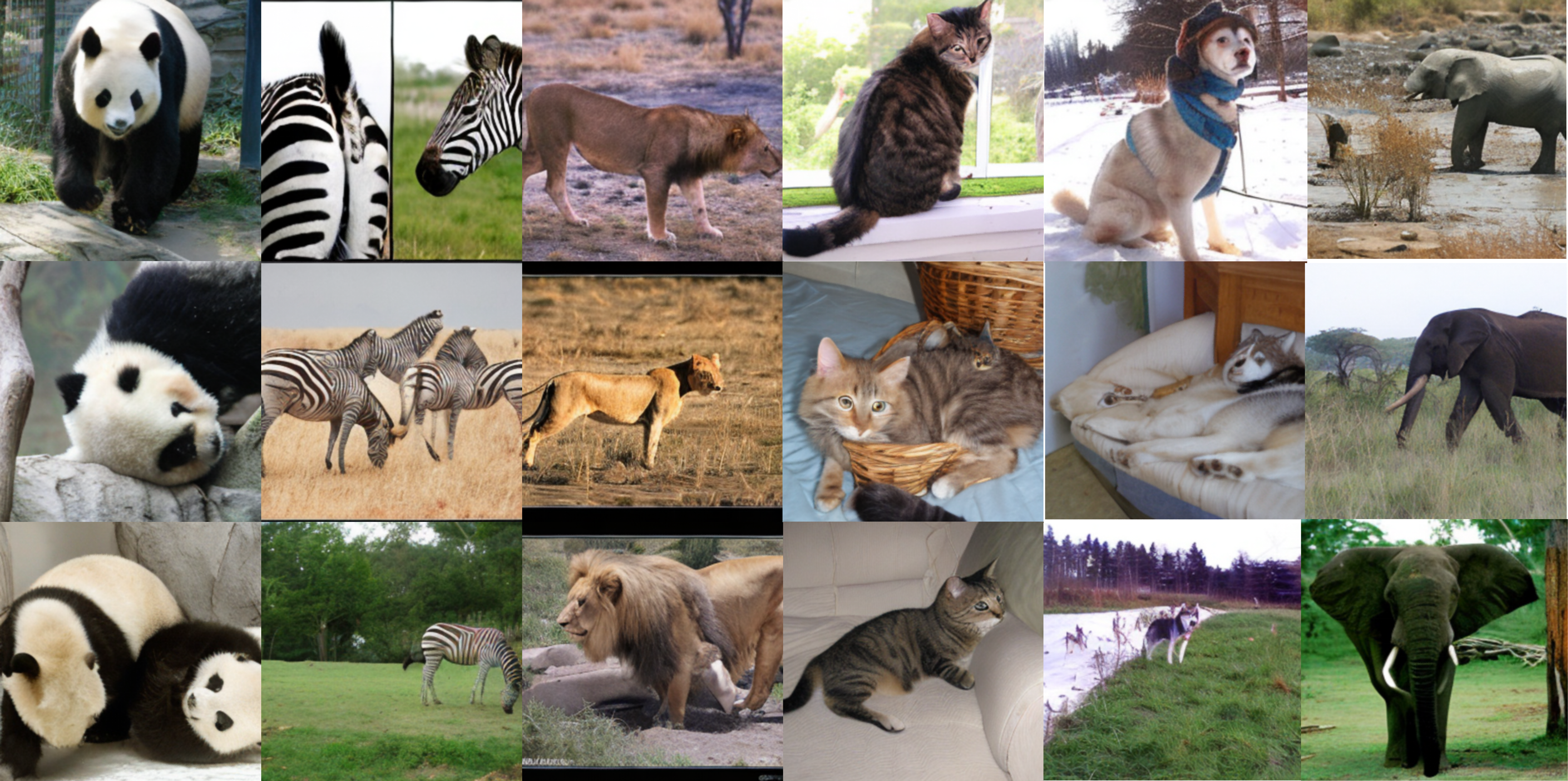}
{\footnotesize
\captionof{figure}{Generation quality of different noise samples.}
\label{fig:noise-sensitivity}
}

\vspace{0.35em}

\includegraphics[width=\textwidth,height=0.41\textheight,keepaspectratio]{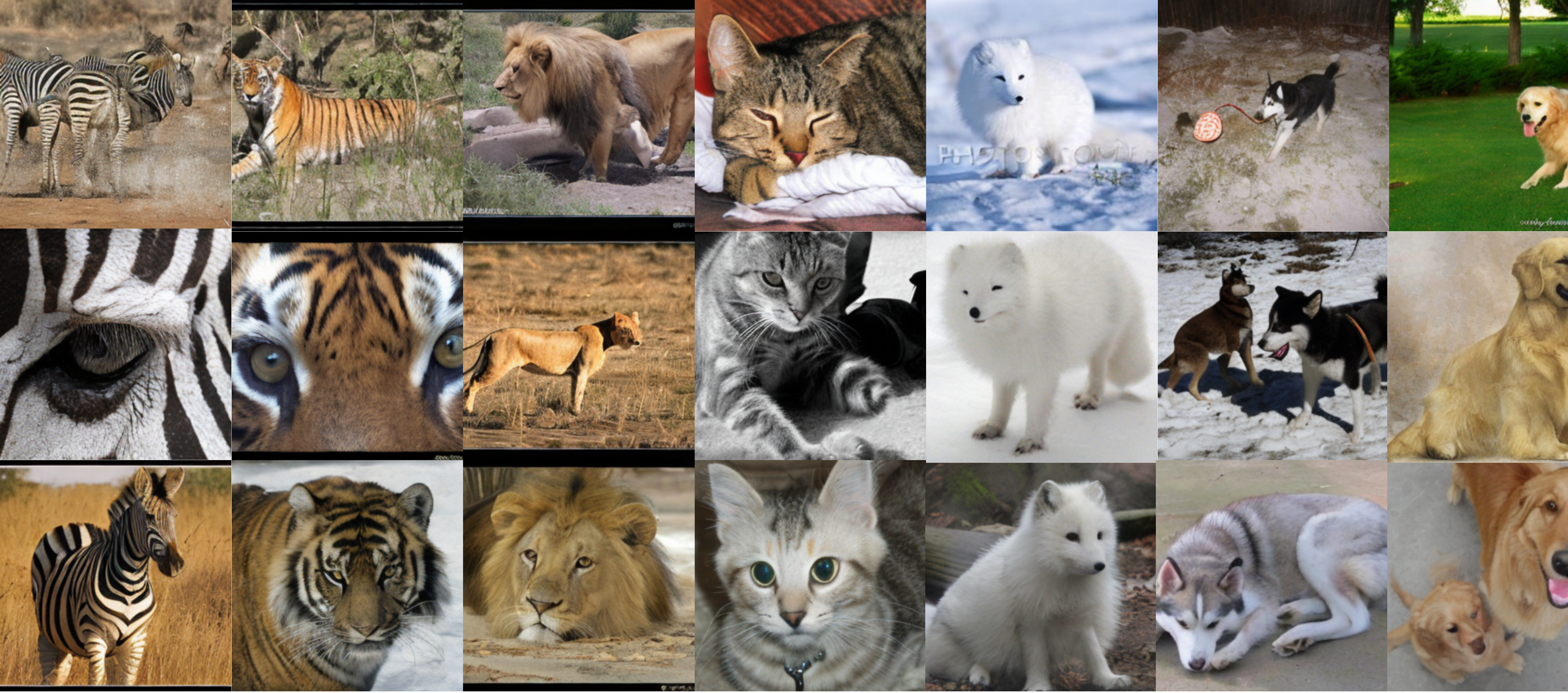}
{\footnotesize
\captionof{figure}{Qualitative comparison between FreeFlow random sampling and our Scout-and-Refine strategy. Each column represents a different ImageNet class (zebra, tiger, lion, tabby cat, Arctic fox, Siberian husky, golden retriever). Row 1--2: Two independent samples from FreeFlow with random initial noise. Row 3: Our method, which uses SLT to evaluate 100 candidate noises and selects the best one for FreeFlow generation.}
\label{fig:qualitative}
}
\end{figure*}

One-step generation models are highly sensitive to initial noise---small differences in noise samples can lead to significant fluctuations in the quality of generation results. Figure~\ref{fig:noise-sensitivity} intuitively demonstrates this phenomenon. It is specifically manifested as structural distortion and semantic conflict of multi-category targets. For example, several generated samples exhibit typical artifacts that reflect sensitivity to initial noise. Notable defects include misaligned facial features in the panda at row 2, column 1, broken stripe textures in the zebra at row 1, column 2, and imbalanced body proportions in the lion at row 1, column 4. Additional issues such as cross-category feature blending and redundant structural elements are also observed in other samples.

Under the same model and settings, changing only the random seed results in significant differences in the output quality of the teacher model, making purely random sampling unreliable. This noise-driven instability directly impairs the reliability and controllability of one-step generation models in practical applications. To address this, we design an inference strategy based on Scout-and-Refine, we use the lightweight model SLT to quickly evaluate and screen a large set of candidate noises, then perform refinement generation only on the optimal noise.Our results are clearly demonstrated in Figure~\ref{fig:qualitative}. The first two rows show one-step generation results from FreeFlow with random initial noise, while the third row presents results after applying our Scout-and-Refine strategy with best-of-100 selection. The random sampling baseline (rows 1--2) exhibits inconsistent quality across different noise seeds, with typical failure cases including: incomplete object structure (the zebra in row 1 is partially cropped), semantic incoherence (row 2 generates a dog-like animal instead of a lion), and visual artifacts (the husky in row 2 shows duplicated dogs in the same frame). In contrast, our Scout-and-Refine results (row 3) consistently produce high-quality images across all categories: complete body structures, well-defined facial features, natural poses, and clear background integration. These improvements demonstrate that our noise screening strategy effectively filters out low-quality initial points.

Table~\ref{tab:time-overhead} compares the time costs of different inference strategies, further demonstrating the feasibility of our scheme. FreeFlow requires 94.60 ms to generate two images from random noise. Evaluating 100 candidate noises with SLT takes 55.54 ms, while refining the selected optimal noise with FreeFlow requires 46.03 ms. Consequently, the candidate evaluation time using SLT is only about 58.7\% of the time needed for FreeFlow to generate two images. The total time of our method is 101.56 ms, which is comparable to the time for FreeFlow to produce two random images, yet it enables Best-of-100 screening to markedly reduce the likelihood of low-quality noise and enhance generation quality.

\begin{table}[H]
\caption{Time overhead comparison of different strategies.}
\label{tab:time-overhead}
\centering
\small
\setlength{\tabcolsep}{3pt}
\renewcommand{\arraystretch}{1.2}
\begin{tabular}{ccc}
\toprule
\makecell[c]{Strategy} & \makecell[c]{Avg.\\time (ms)} & \makecell[c]{Std.\\dev. (ms)} \\
\midrule
FreeFlow (2) & 94.60 & 44.49 \\
SLT ($N=100$) & 55.54 & 10.11 \\
FreeFlow Refine (1) & 46.03 & 12.53 \\
Ours (total) & 101.56 & 16.55 \\
\bottomrule
\end{tabular}
\end{table}

 \section{Conclusion}
 Building on the one-step formulation of FreeFlow, we first derive a depth-wise ODE and reinterpret the multi-layer Transformer stack as its Euler-discretized process. This allows us to distill the teacher’s layered transformations into a single shared-weight block, yielding the student model SLT, which is a single-layer network that aligns each forward step with a distinct depth in the teacher. As a result, SLT compresses FreeFlow’s 28 layers into a single layer, achieving a 155× reduction in parameters. Second, we exploit SLT’s efficiency via a Scout-and-Refine strategy, screening over 100 noise seeds to select an optimal starting point for final refinement with FreeFlow. Experiments show that this co-design markedly improves output quality by sidestepping low-quality noise. In summary, beyond reducing sampling steps, our method establishes network depth compression, as exemplified by distilling FreeFlow into a discretized single-layer student, as a powerful new axis for efficient diffusion inference.

\section*{Acknowledgements}
This work was supported in part by the National Natural Science Foundation of China under Grant 62271504, and by the Key Research and Development and Promotion Project of Henan Province under Grant 251111312900.

\clearpage

\FloatBarrier

\clearpage
 \bibliography{references}
 \bibliographystyle{icml2026}
 \end{document}